\documentclass[letterpaper, 10 pt, conference]{ieeeconf}  

\IEEEoverridecommandlockouts                              

\usepackage{graphicx}
\usepackage{subcaption}
\usepackage[font=small,labelfont=bf]{caption}
\usepackage{mwe}
\usepackage{amsmath}
\usepackage{tablefootnote}
\usepackage{url}
\usepackage{listings}
\usepackage{float}
\usepackage{multirow}
\usepackage{hhline} 
\usepackage{tablefootnote}
\usepackage{dingbat}
 \usepackage[table]{xcolor}
\usepackage[dvipsnames]{xcolor}
\usepackage{tikz}
\usepackage{bbding}
\usepackage{color}
 \usepackage[table]{xcolor}
\usepackage{cuted}
\usepackage{capt-of}
\usepackage{makecell}
\definecolor{cvprblue}{rgb}{0.21,0.49,0.74}
\usepackage{arydshln}
\usepackage[pagebackref,breaklinks,colorlinks,citecolor=cvprblue]
{hyperref}
\usepackage{array} 

\usepackage{booktabs}
\usepackage{colortbl}
\usepackage{xcolor}
\usepackage{pifont}
\usepackage{graphicx}
\usepackage{subcaption}

\newcommand{\cmark}{\textcolor{green!60!black}{\ding{51}}}
\newcommand{\xmark}{\textcolor{red}{\ding{55}}}

\title{\LARGE \bf
\textsc{MOTOR}: A Multimodal Dataset for Two-Wheeler Rider Behavior Understanding
}

\author{
    Varun A. Paturkar$^{1}$, Shankar Gangisetty$^{1}$, C.V. Jawahar$^{1}$%
    \thanks{$^{1}$CVIT, IIIT-Hyderabad, India.
    {\tt\small \{varuna.paturkar@research., shankar.gangisetty@ihub-data., jawahar@\}iiit.ac.in}}%
}

\begin{document}
\maketitle

\begin{abstract}
Two-wheelers account for a disproportionately high share of road fatalities in the Global South. Research on two-wheeler rider behavior, however, lags far behind four-wheelers, where multimodal datasets have driven major advances in Advanced Driver Assistance Systems (ADAS). To address this gap, we present the MOtorized TwO-wheeler Rider (MOTOR) dataset, the first large-scale, multi-view, multimodal resource dedicated to two-wheelers in dense, unstructured traffic. MOTOR comprises 1,629 sequences (25+ hours of video data) collected from 16 riders and integrates synchronized front, rear, and helmet videos, rider eye-gaze from wearable trackers, on-road audio, and telemetry (GPS, accelerometer, gyroscope). Rich annotations capture traffic context, rider state, 12 riding maneuvers spanning conventional and unconventional behaviors, and legality labels (\textit{Legal}, \textit{Illegal}, \textit{Unspecified}). We benchmark rider behavior recognition and maneuver legality classification using state-of-the-art video action recognition backbones (CNN and Transformer-based), extended with multimodal fusion, and find that combining RGB, gaze, and telemetry consistently yields the best performance. MOTOR thus provides a unique foundation for advancing safety-critical understanding of two-wheeler riding. It offers the research community a benchmark to develop and evaluate models for behavior analysis, legality-aware prediction, and intelligent transportation systems. Dataset and code is available at \mbox{\url{https://varuniiith.github.io/MOTOR-Dataset/}}
\end{abstract}

\section{Introduction}
\label{Sec_Introduction}
\begin{figure}[t]
    \centering
    \includegraphics[width=\columnwidth]{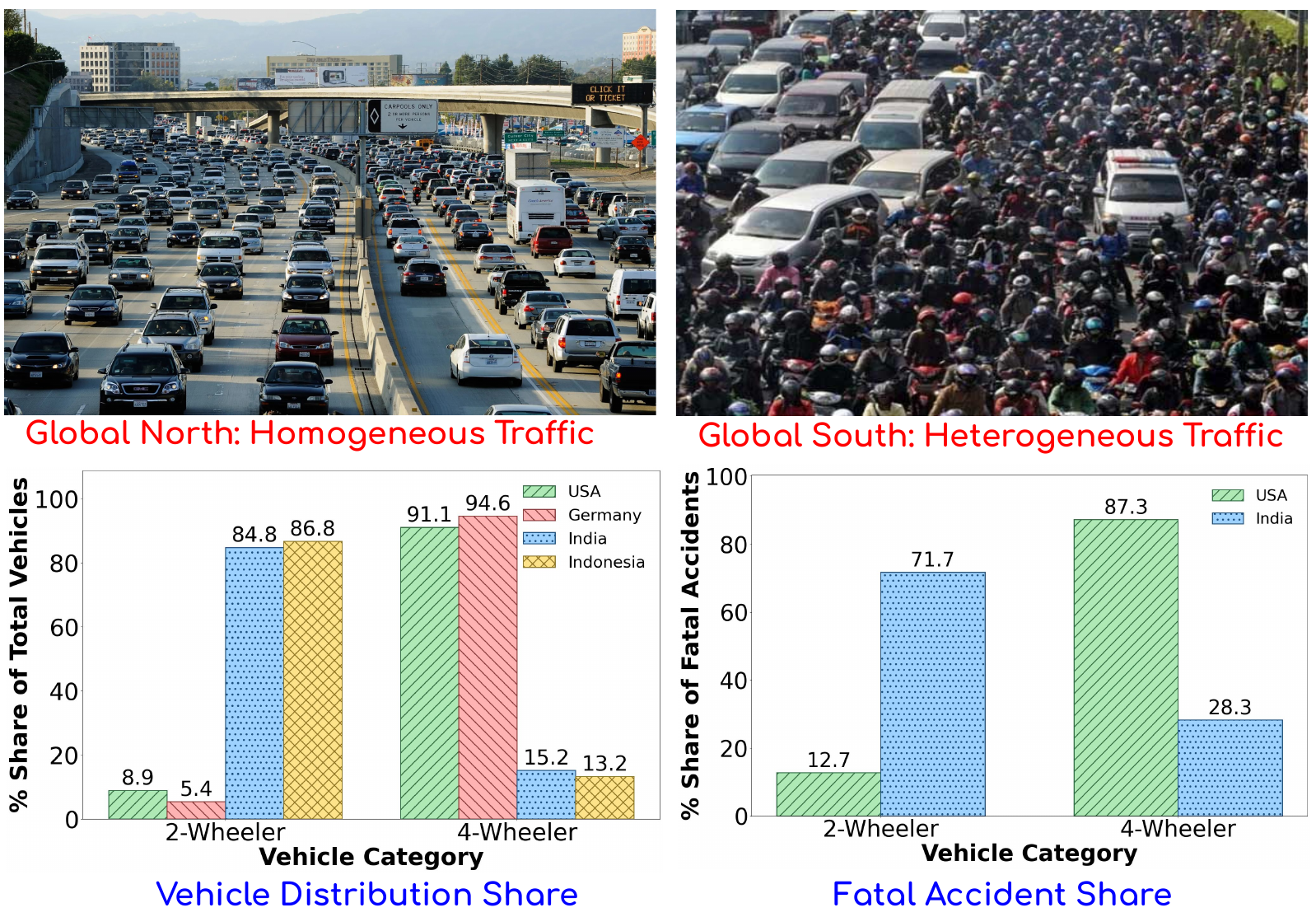}
    \caption{\textbf{Comparison of traffic contexts and accident statistics across the Global North and South.} \textit{Top Row:} Four-wheelers dominating in the USA vs Two-wheelers in India. \textit{Bottom Row:} Distribution of vehicles (two-wheeler vs four-wheeler) and fatal accidents across North and South.
    }
    \label{fig:global_nvs_s}
\end{figure}

\noindent Four-wheelers dominate road transport in the Global North, with cars being the primary mode of transport in regions such as the USA~\cite{cars_usa} and Germany~\cite{cars_germany}. In contrast, motorized two-wheelers (motorbikes and scooters) are the predominant means of transport in the Global South, including countries like India~\cite{morth} and Indonesia~\cite{indonesia}. This asymmetry has strongly influenced research priorities and safety technologies. 
The availability of large-scale multimodal driving datasets, including modalities such as RGB videos, LiDAR, GPS/IMU, vehicle telemetry and driver gaze has fueled breakthroughs in four-wheeler tasks such as object detection~\cite{caesar2020nuscenes,sun2020scalability,li2023large,yu2020bdd100k}, semantic segmentation~\cite{varma2019idd,huang2018apolloscape}, intention prediction~\cite{ramanishka2018toward,jain2016brain4cars}, and trajectory forecasting~\cite{chandra2023meteor,wasi2024early, mahjourian2024unigen}. Despite constituting a larger share of on-road vehicles and fatal accidents in the Global South~\cite{morth_accidents},  two-wheeler behavior remains underexplored, limiting the development of safety-critical models. 
\noindent As shown in Fig.~\ref{fig:global_nvs_s}, two-wheelers are the dominant mode of commute in the Global South, central to commercial activities such as delivery services and ride-sharing. Their widespread use raises critical safety concerns: unlike cars, two-wheelers are more prone to accidents due to \textit{sudden acceleration and braking}, \textit{significant lean angles during maneuvers}, and \textit{minimal structural protection}. These dynamics, coupled with close interactions in dense traffic, introduce risks rarely captured in four-wheeler datasets. This gap underscores the need for dedicated resources to study two-wheeler riding behaviors, a challenge our work directly addresses.


\begin{figure*}[t] 
  \centering
  \includegraphics[width=\textwidth]{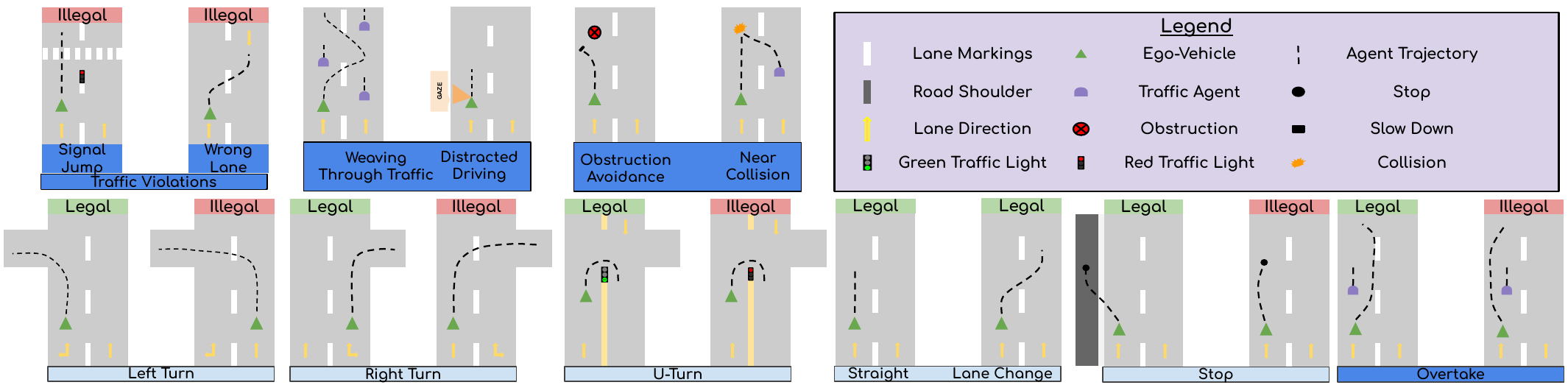}
  \caption{\textbf{Illustration of rider behaviors in the MOTOR dataset.} Light blue indicates conventional behaviors, dark blue indicates unconventional behaviors. Legal and illegal maneuvers are shown where applicable; behaviors without explicit legality labels are marked as unspecified.}
  \label{fig:rider_behaviours}
\end{figure*}

\noindent Recent efforts have begun to address two-wheeler behavior through dedicated datasets. The most notable is RAAD~\cite{gangisetty2024icpr}, which studied rider intention prediction; myEye2Wheeler~\cite{kumar2024myeye2wheeler} introduced rider eye-tracking. As shown in Table~\ref{tab:dataset_comparison}, these datasets are limited in accessibility, usability, details, size, number of views, modalities, conventional, and unconventional riding behaviors.



\noindent To address these gaps, we present the MOtorized TwO-wheeler Rider (MOTOR) dataset, the first multi rider, multi-view, multimodal dataset designed specifically for two-wheelers. MOTOR comprises 1,629 sequences collected from 16 riders, integrating synchronized multi-view videos (ego-vehicle front, rear, and helmet views), rider eye-gaze from wearable trackers (Aria~\cite{engel2023projectarianewtool}, Pupil~\cite{kassner2014pupilopensourceplatform}), on-road audio capturing ambient sounds and distractions, and telemetry signals (GPS, accelerometer, gyroscope) for speed, location, and lean angles. The dataset is richly annotated with traffic scene context, rider state, six conventional and six unconventional behaviors, along with their legality labels (legal or illegal). As illustrated in Fig.~\ref{fig:rider_behaviours}, MOTOR  captures how riders behave and whether these behaviors comply with traffic rules, offering a unique foundation for safety-critical analysis in two-wheeler research. We benchmark rider behavior recognition on the MOTOR dataset using state-of-the-art action recognition backbones: CNN-based S3D~\cite{xie2018rethinkingspatiotemporalfeaturelearning}, ResNet3D~\cite{tran2018closerlookspatiotemporalconvolutions}, and Transformer-based Video Swin Transformer~\cite{liu2021videoswintransformer}, MViTv2~\cite{li2022mvitv2improvedmultiscalevision}, extended with multimodal fusion. Beyond behavior classification, we also predict maneuver legality (Legal, Illegal, Unspecified), enabling a more comprehensive understanding of rider actions through visual and non-visual cues.


\noindent The main contributions of our work are:
\begin{itemize}
\item We introduce \textbf{MOTOR}, the first large-scale, multi-rider, multi-view, multimodal dataset for two-wheelers in dense and unstructured traffic. The dataset comprises 1,629 sequences (25+ hours of video data) from 16 riders with synchronized front, rear, and helmet videos, rider eye-gaze, on-road audio, and telemetry. It captures diverse two-wheeler behaviors (see Fig.~\ref{fig:rider_behaviours}, Fig.~\ref{fig:critical_manoeuvres}) including near-collisions, traffic violations, distracted driving, and interactions with vehicles, pedestrians, and street vendors.

\item We provide rich annotations of traffic context, rider state, six conventional and six unconventional riding maneuvers, together with legality labels (\textit{Legal}, \textit{Illegal}, \textit{Unspecified}), enabling a legality-aware analysis of how riders behave and comply with traffic rules.

\item We benchmark rider behavior recognition and maneuver legality classification using state-of-the-art video action recognition backbones, extended with multimodal fusion of gaze and telemetry, showing that combining modalities consistently improves performance across both tasks compared to video-only baselines.

\item We conduct exhaustive experiments to analyze the contribution of each modality (video, gaze, telemetry) and further provide a detailed class-wise accuracy analysis across all backbones (see Fig~\ref{fig:confusion_matrices}), offering deeper insights into two-wheeler behavior understanding.

\end{itemize}


\section{Related Works}
\label{related-works}

\begin{table*}[t]
\centering
\caption{\textbf{Comparison of 4\text{-}wheeler and 2\text{-}wheeler behavior datasets.}  
Our dataset is unique as it contains multi-modal, multi-view videos from ego-vehicle and helmet, eye gaze, as well as annotated conventional and unconventional behaviors, and legality-related riding scenarios.  
Note: CRB indicates conventional riding behaviors, and UCRB means unconventional riding behaviors.}
\resizebox{\textwidth}{!}{
\begin{tabular}{lcccccccccccc}
\toprule
\textbf{Dataset} & \textbf{Ego-vehicle} & \textbf{\#Clips} & \textbf{\makecell{Duration \\(hrs)}} & \textbf{\#Views} & \textbf{Resolution} & \textbf{Multi-Modal} & \textbf{\makecell{Telemetry \\ (gyro,accelero)}} & \textbf{Eye Gaze} & \textbf{Audio} & \textbf{CRB} & \textbf{UCRB} & \textbf{\makecell{Legality\\(legal,illegal)}} \\
\midrule
IDD\text{-}X~\cite{parikh2024idd}      & Car & 3{,}635   & 85    & 2 & 2560$\times$1440 & \xmark & \xmark & \xmark & \xmark & \cmark & Partial  & \xmark \\
HDD~\cite{ramanishka2018toward}        & Car & 137       & 104   & 3 & 1920$\times$1200 & \cmark & \cmark & \xmark & \xmark & \cmark & Partial & \xmark \\
DAAD~\cite{wasi2024early}              & Car & 2{,}028   & 85    & 6 & 1920$\times$1080 & \cmark & \xmark & \cmark & \xmark & \cmark & \xmark  & \xmark \\
Brain4Cars~\cite{jain2016brain4cars}   & Car & --        & 10    & 2 & --              & \cmark & \cmark & \xmark & \xmark & \cmark & \xmark  & \xmark \\
METEOR~\cite{chandra2023meteor}        & Car & 1{,}250   & --    & 2 & 1920$\times$1080 & \cmark & \xmark & \xmark & \xmark & \cmark & Partial & \cmark \\
AIDE~\cite{yang2023aide}               & Car & --        & 2.4   & 4 & 1920$\times$1080 & \cmark & \xmark & \xmark & \xmark & \cmark & \xmark  & \xmark \\

\hdashline

RAAD~\cite{gangisetty2024icpr}               & 2\text{-}Wheeler & 1{,}000   & --    & 3 & 1920$\times$1080 & \xmark & \xmark & \xmark & \xmark & \cmark & \xmark  & \xmark \\
myEye2Wheeler~\cite{kumar2024myeye2wheeler} & 2\text{-}Wheeler & --        & 100+  & 1 & 1920$\times$1080 & \cmark & \cmark & \cmark & \xmark & \xmark & \xmark  & \xmark \\
Oxford RobotCycle~\cite{panagiotaki2025oxford} & 2\text{-}Wheeler & -- & -- & 4 & -- & \cmark & \cmark & \xmark & \xmark & \cmark & \xmark & \xmark \\
\textbf{MOTOR (Ours)}                       & \textbf{2\text{-}Wheeler} & \textbf{1{,}629} & \textbf{25} & \textbf{4} & \textbf{1920$\times$1080} & \textbf{\cmark} & \textbf{\cmark} & \textbf{\cmark} & \textbf{\cmark} & \textbf{\cmark} & \textbf{\cmark} & \textbf{\cmark} \\
\bottomrule
\end{tabular}
}

\label{tab:dataset_comparison}
\end{table*}

\noindent We summarize existing driver (Four-Wheeler) and rider (Two-Wheeler) behavior datasets in Table~\ref{tab:dataset_comparison}.  


\subsection{Four-Wheeler Driver Behavior}
\noindent Driver behavior has been extensively studied in the context of cars through a wide range of datasets~\cite{jain2016brain4cars, ramanishka2018toward, yang2023aide,chandra2023meteor, parikh2024idd}. Brain4Cars~\cite{jain2016brain4cars} introduced driver and road-facing videos for maneuver recognition, while HDD~\cite{ramanishka2018toward} extended this with hierarchical annotations of naturalistic driving. The DMD~\cite{ortega2020dmd} dataset focused more on driver alertness, whereas AIDE~\cite{yang2023aide} captured driver behavior from both inside and outside the vehicle. However, these datasets were primarily collected in structured and sparse traffic. More recent works such as METEOR~\cite{chandra2023meteor} and IDD-X~\cite{parikh2024idd} expanded coverage to heterogeneous and unconstrained road conditions. In parallel, gaze-focused datasets such as DR(eye)VE~\cite{palazzi2018predicting}, LBW~\cite{kasahara2022look}, and DAAD~\cite{wasi2024early} highlighted the importance of attention cues in driving. Together, these efforts showcase the richness and diversity of four-wheeler datasets across structured, unstructured, and gaze-aware contexts.

\noindent Nonetheless, they remain centered on cars, with in and out-cabin mounted viewpoints and relatively stable vehicle dynamics. Such setups fail to capture the unique challenges of two-wheelers, including high lean angles during turns, rapid acceleration and braking, and frequent behaviors such as weaving through dense traffic. These limitations underscore the need for dedicated rider-focused datasets.

\subsection{Two-Wheeler Tasks, Datasets, and Methods}  
\noindent Recently, there has been a growing interest in the two-wheeler space, with tasks such as riding pattern recognition~\cite{6899632}, riding dynamics analysis~\cite{10122165}, safety gear recognition~\cite{10717752}, as well as datasets~\cite{gangisetty2024icpr,8693888,figueiredo2021more,panagiotaki2025oxford}, yet the datasets are limited. The RAAD dataset~\cite{gangisetty2024icpr} studied rider intention prediction using short clips of six conventional riding maneuvers. While valuable as a first step, RAAD captures only the rider and surrounding vehicles, without incorporating rider gaze. Similarly, myEye2Wheeler~\cite{kumar2024myeye2wheeler} introduced an egocentric rider gaze dataset, but its scope is restricted to attention modeling and does not extend to behavioral tasks. Other two-wheeler datasets and efforts remain task-specific: CDBV~\cite{8693888} collected bike-mounted egocentric views, MoRe~\cite{figueiredo2021more} targeted motorcycle re-identification, while riding dynamics~\cite{10122165} and powered two-wheeler patterns~\cite{6899632} explored specific motion and traffic patterns. The Oxford RobotCycle~\cite{panagiotaki2025oxford} examined robotic two-wheeler control, while mobility studies such as EMBARQ~\cite{EMBARQ} highlighted the prevalence of motorized two-wheelers in urban settings. Collectively, these works indicate a growing interest in two-wheeler research but remain narrow in scope. In contrast, our MOTOR dataset introduces a multi-rider, multi-view  and multimodal collection with rich annotations covering both conventional and unconventional behaviors along with their legality. This level of diversity opens up a broad research space for advancing rider safety and developing robust models for two-wheeler scenarios.


\begin{figure}[t]
    \centering
    \includegraphics[width=\columnwidth]{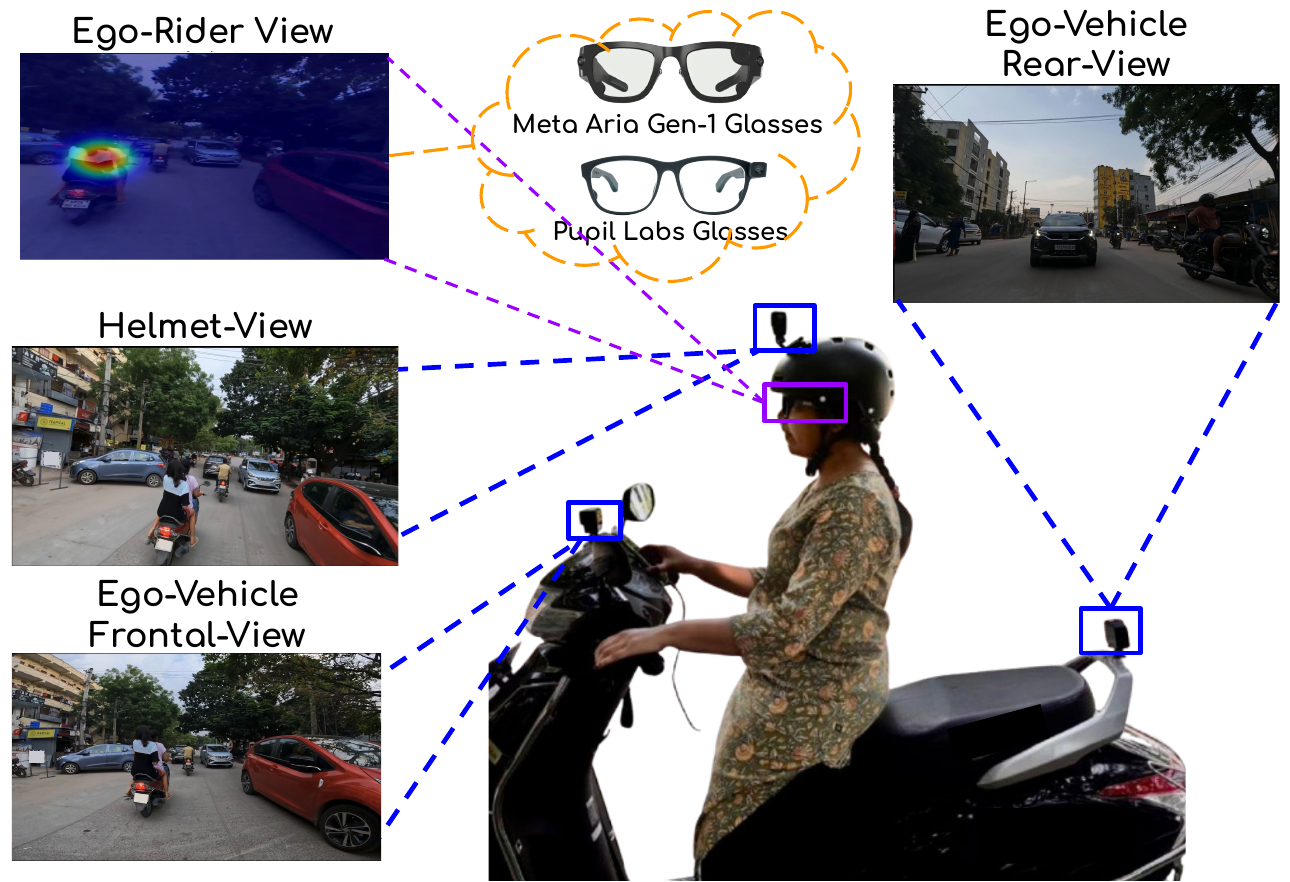} 
    \caption{\textbf{Data capture setup}: Three cameras are oriented towards front-view (ego-vehicle, helmet-mounted), rare-view, and eye-gaze derived from eye-tracking cameras (Aria~\cite{engel2023projectarianewtool} or Pupil~\cite{kassner2014pupilopensourceplatform} glasses).}
    \label{fig:data_capture_setup}
\end{figure}

\section{The MOTOR Dataset}
\noindent In this section, we introduce our dataset and present details of data collection, annotation, and data statistics.

\subsection{Data Collection}
\label{data-capture}

\begin{figure*}[t]
    \centering
    \includegraphics[width=\textwidth]{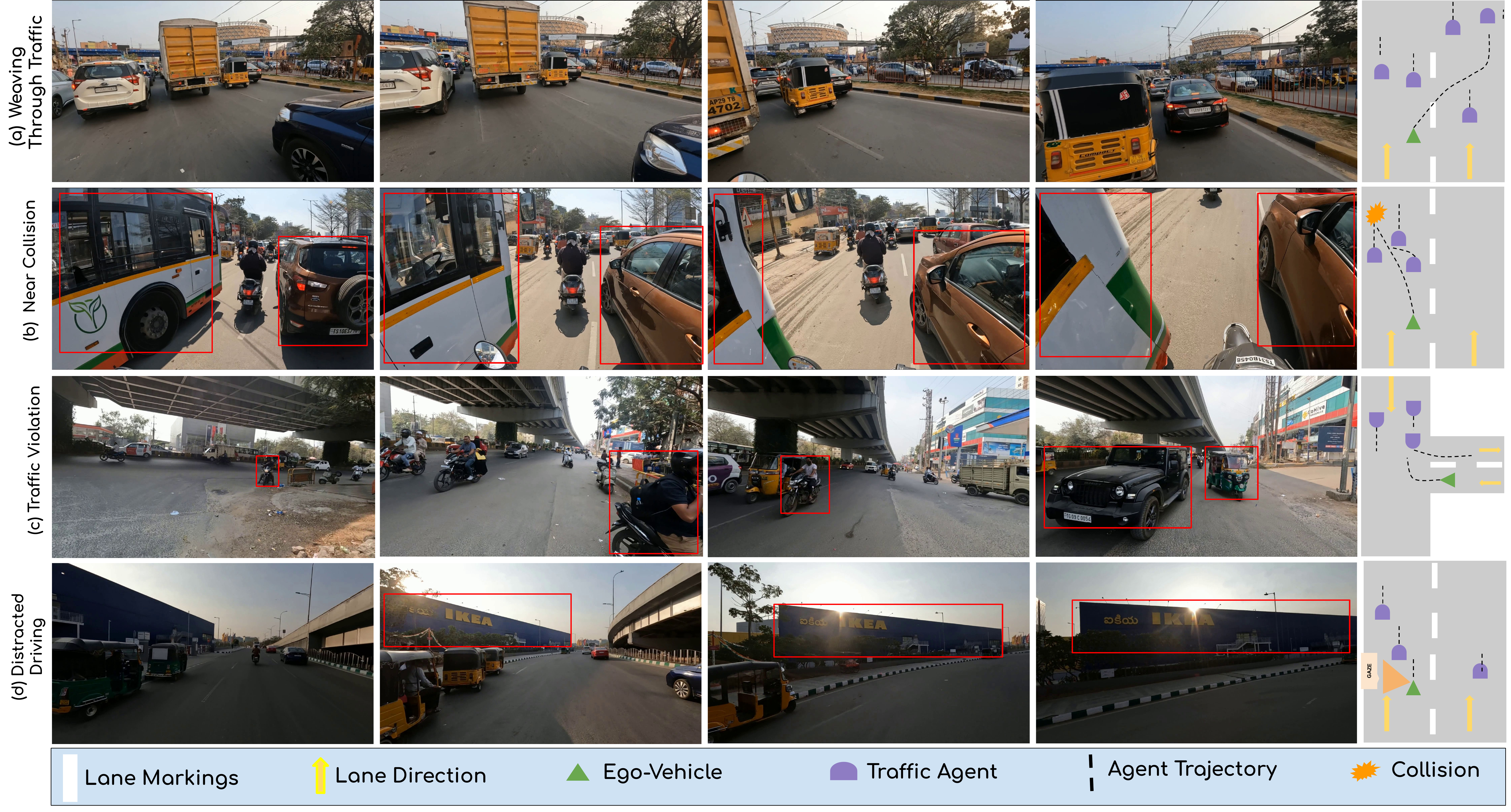}
    \caption{\textbf{Data samples helmet-view.} 
    \textit{(a)} Ego-rider weaves through dense, slow traffic, overtaking multiple vehicles across lanes.  
    \textit{(b)} Rider squeezes through a narrow gap between a bus and a car, narrowly avoiding the bus.  
    \textit{(c)} Rider rides in the wrong lane against dense oncoming traffic, disrupting flow.  
    \textit{(d)} Rider turns head fully toward a roadside building, diverting gaze from the road amid fast-moving traffic.}
    \label{fig:critical_manoeuvres}
\end{figure*}

\noindent \textbf{Data Capture Platform.} 
The MOTOR dataset was collected using a multi-modal multi-view setup as shown in Fig.~\ref{fig:data_capture_setup} with three GoPro-10 cameras and wearable eye-tracking glasses. A front-mounted GoPro, a helmet-mounted GoPro, and a rear-mounted  GoPro on the grab rail. Additionally, eye-tracking glasses (Project Aria~\cite{engel2023projectarianewtool} or Pupil Labs~\cite{kassner2014pupilopensourceplatform}) captured the rider’s egocentric view along with gaze information. MOTOR comprises multiple data streams that are synchronized and timestamped.


\noindent \textbf{Video Data}. The vehicle was equipped with three GoPro Hero 10 cameras with 1920$\times$1080 resolution, 30 FPS, and video stabilization enabled. The front and rear cameras captured the ego-vehicle’s frontal and rear views, providing surrounding traffic context, including side blind spots. The helmet-mounted camera recorded the rider’s egocentric view, reflecting the rider's visual perspective of the environment and head movements relevant for rider behavior.

\noindent \textbf{Rider Gaze}. Gaze data was recorded using either Aria~\cite{engel2023projectarianewtool} or Pupil~\cite{kassner2014pupilopensourceplatform} eye-tracking cameras. The Aria device provides eye-tracking at 320$\times$240 resolution and includes an 8 MP RGB camera recording at 1408$\times$1408. In contrast, the Pupil device offers eye-tracking at 192$\times$192 resolution with a 200 Hz sampling rate and a scene camera that captures the rider’s egocentric view at 1088$\times$1080. 

\noindent \textbf{Audio and Telemetry Data}. The Aria~\cite{engel2023projectarianewtool}, Pupil~\cite{kassner2014pupilopensourceplatform} devices, and the GoPro cameras were also used to capture ambient traffic sounds such as horns, engine noise, and rider speech (e.g., with a pillion or during phone calls), which may contribute to riding distractions. Telemetry data was extracted from the GoPro recordings, including GPS signals sampled at 10 Hz for location tracking and inertial measurements (accelerometer and gyroscope) sampled at 200 Hz to characterize rider behavior and ego-vehicle dynamics. This telemetry was synchronized with the video streams using the GoPro telemetry extractor~\cite{telemetry_extractor}.


\noindent \textbf{Data Collection.} 
The dataset comprises 25 hours of riding data collected over 4 weeks, consisting of 25 unique sequences recorded from 16 riders (13 male and 3 female). The diversity of riding data includes varying traffic densities (from peak-hour congestion to sparse early-morning traffic), a wide range of rider experience (2 to 20 years), multiple road types (paved and unpaved, with and without lane markings), and different two-wheeler vehicle types. Importantly, the dataset contains several instances of ego-rider traffic violations, near-collision events, and interactions with pedestrians, street vendors, potholes, and traffic barricades. A comparison of MOTOR with existing two-wheeler and four-wheeler driving datasets is provided in Table~\ref{tab:dataset_comparison}, while Fig.~\ref{fig:critical_manoeuvres} illustrates naturalistic unstructured riding maneuvers leading to critical situations.

\subsection{Data Annotation and Statistics}
\label{data-feature}
\noindent To capture a comprehensive view of the traffic scene, ego-rider state, and rider behaviors, the dataset annotations are categorized into four types.

\noindent \textbf{(i) Traffic Scene and Rider State Annotations}.

\noindent \textit{Traffic scene annotations} capture the context of ego-rider operation, including time of day, road surface (paved/unpaved), number of lanes, presence of lane markings or dividers, and traffic density. Whereas, \textit{Ego rider state annotations} capture the state of the rider and vehicle, including GPS trajectories, 2D vehicle speeds, rider gaze points, and overall gaze behavior during maneuvers. Gaze behavior is categorized into four classes: \textit{looking straight ahead} (LS), \textit{looking right} (LR), \textit{looking left} (LL), and \textit{glancing sideways} (GS).

\noindent \textbf{(ii) Conventional Rider Behaviors.}
These annotations capture standard maneuvers performed by the ego-rider, including \textit{Going Straight} (GS), \textit{Left Turn} (LT), \textit{Right Turn} (RT), \textit{Lane Change} (LC), \textit{U-Turn} (UT), and \textit{Stop}.

\noindent \textbf{(iii) Unconventional Rider Behaviors.}
These annotations capture maneuvers that are potentially dangerous for the rider and should be performed with extreme caution or avoided altogether (see Fig.~\ref{fig:critical_manoeuvres}). We categorize them as follows:
    \begin{itemize}
        \item \textit{Overtaking (OT)}: Ego-rider overtakes another traffic agent by accelerating.
        \item \textit{Weaving Through Traffic (WTT)}: Ego-rider maneuvers through dense traffic by frequently switching between small gaps to move ahead.
        \item \textit{Obstruction Avoidance (OA)}: Ego-rider swerves or slows down to avoid an obstruction. Sub-categories include avoidance of a vehicle, pedestrian, traffic barricade, or pothole/speed breaker.
        \item \textit{Distracted Riding (DD)}: Ego-rider diverts attention away from the road ahead, e.g., by looking at a phone, billboard, or other off-road stimuli.
        \item \textit{Traffic Violations (Vio)}: Instances where the ego-rider commits a violation as defined by the Indian Motor Vehicle Act~\cite{Indian_Motor_Vehicle_Act}. Sub-categories include signal jumps, wrong-lane riding, illegal turns and illegal parking.
        \item \textit{Near Collisions (NC)}: Maneuvers that lead to or narrowly avoid a collision with another traffic agent.
    \end{itemize}

\begin{figure}[t]
    \centering
    \includegraphics[width=\linewidth]{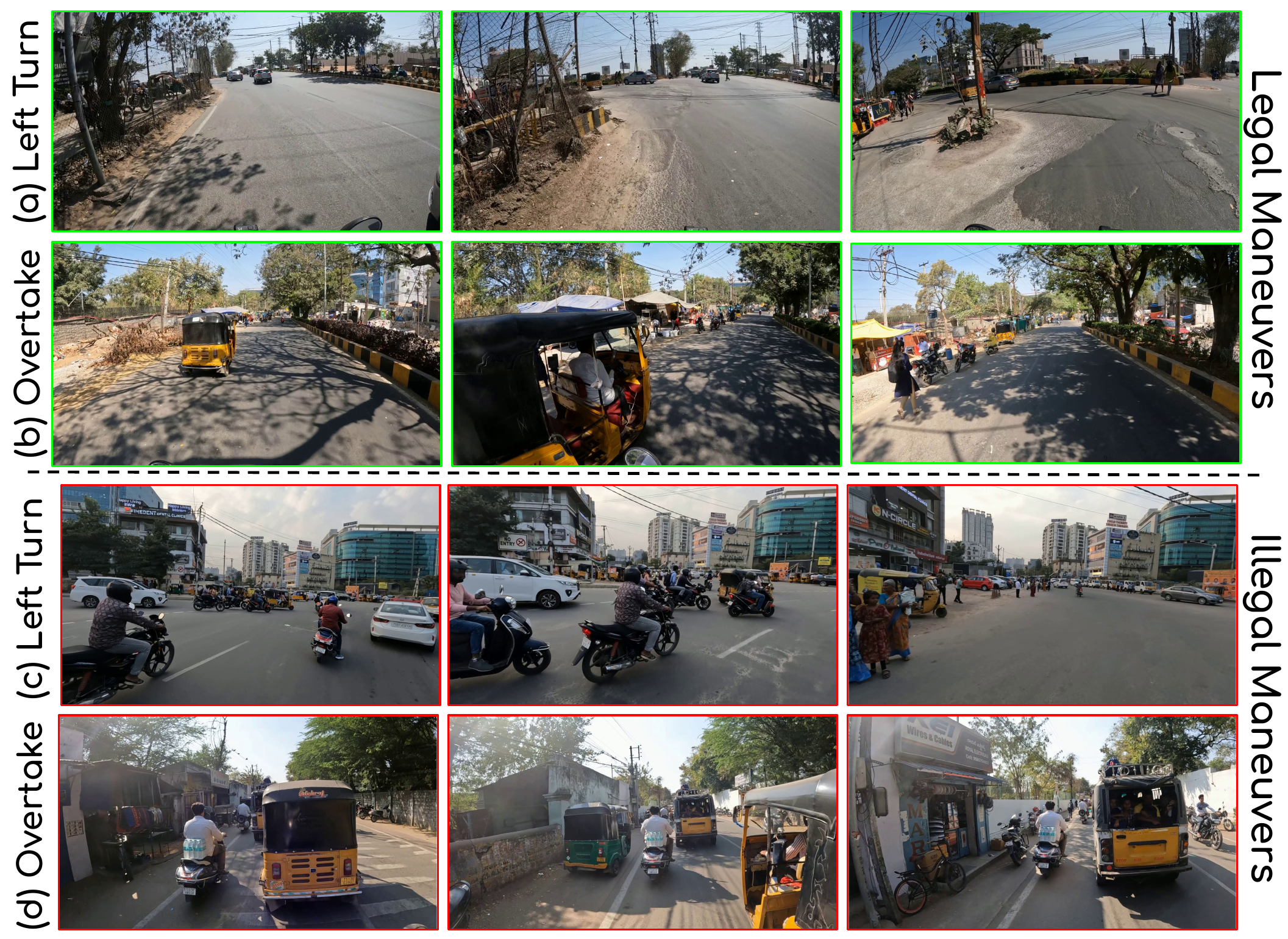} 
    \caption{\textbf{Data samples of legal and illegal maneuvers.} 
    \textit{(a) Legal Left Turn:} The rider correctly takes the left turn by positioning in the left-most lane in advance. \textit{(b) Legal Overtake:} The rider legally overtakes a three-wheeler using the right lane. \textit{(c) Illegal Left Turn:} The rider makes a left turn from the right-most lane, disrupting traffic flow. \textit{(d) Illegal Overtake:} The rider overtakes a three-wheeler from the left within the same lane, even though the right lane is vacant.}
    \label{fig:legal_illegal_examples}
\vspace{-.5cm}
\end{figure}

\begin{figure}[t]
    \centering
    \includegraphics[width=\linewidth]{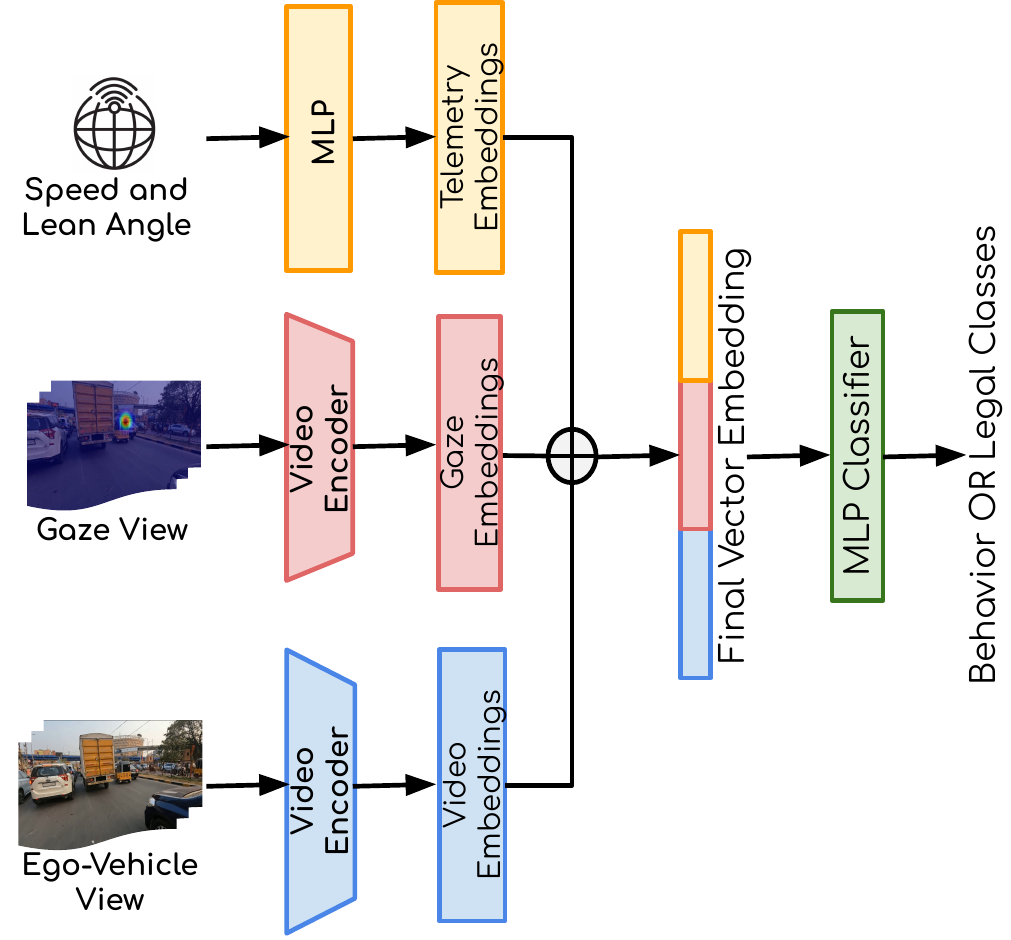} 
    \caption{\textbf{Our proposed baseline for rider behavior and legality classification.} The dual video encoder generates the spatio-temporal features (video and gaze embeddings) along with telemetry features from MLP. These embeddings are concatenated and fed into the MLP classifier to predict the behavioral or legality classes.}
    \label{fig:architecture}
\end{figure}

\noindent \textbf{(iv) Legality Annotations.}
Each maneuver is annotated for its legality based on the Indian Motor Vehicle Act (2017)~\cite{Indian_Motor_Vehicle_Act} (see Fig.~\ref{fig:legal_illegal_examples}). Legality is categorized into three classes:
\begin{itemize}
    \item \textit{Legal}: Maneuvers that comply with traffic rules, such as lane following, legal turns, and stopping at signals.
    \item \textit{Illegal}: Maneuvers that explicitly violate traffic laws, such as signal jumps, wrong-lane riding, illegal turns, or illegal parking.
    \item \textit{Unspecified}: Maneuvers whose legality depends on context and is not explicitly defined in the Act, such as weaving through traffic or obstruction avoidance.  
\end{itemize}

\begin{figure*}[t]
    \centering

    \begin{subfigure}{0.32\textwidth}
        \includegraphics[width=\linewidth,height=4.5cm]{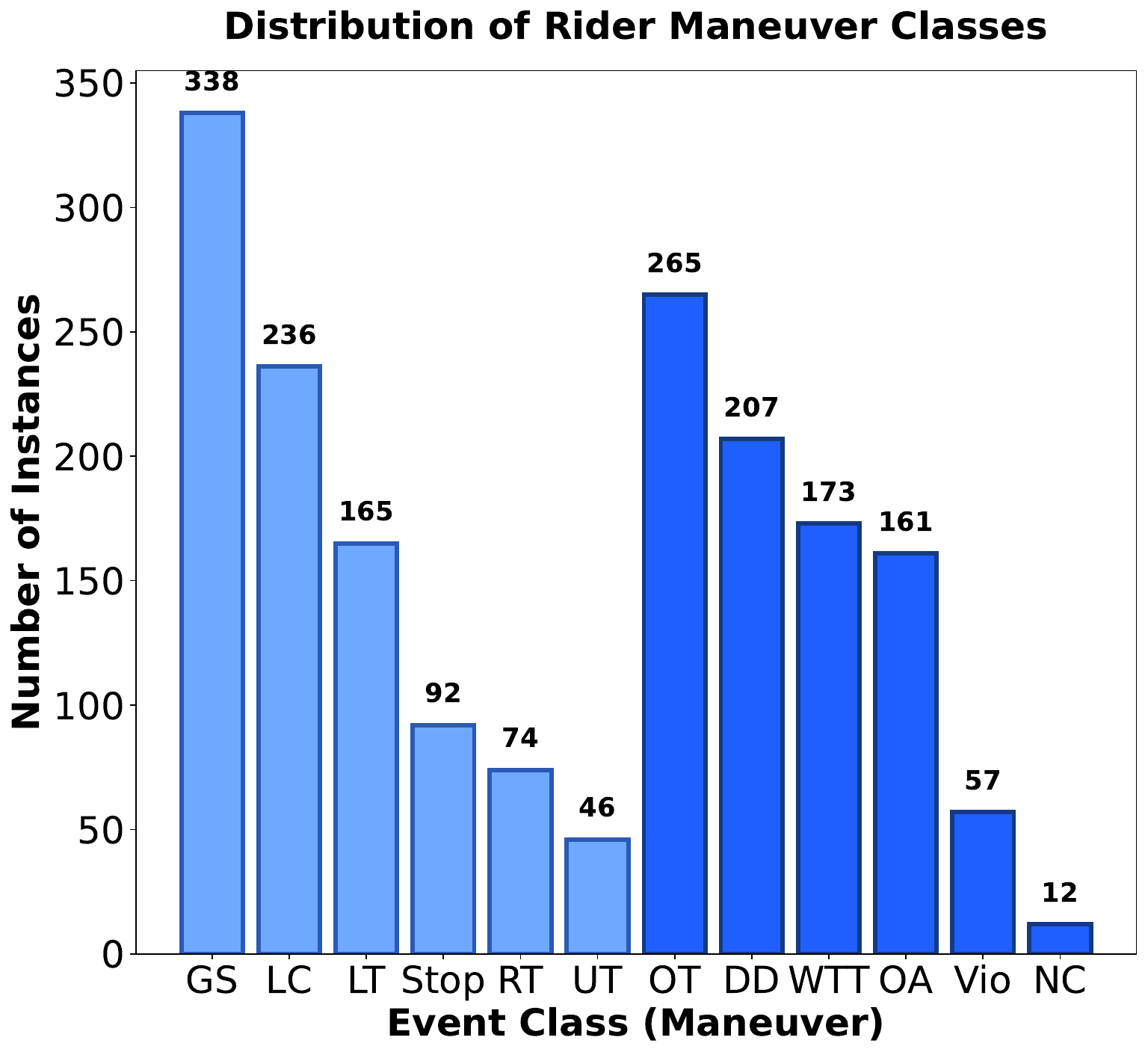}
        \caption{}
        \label{fig:plots_a}
    \end{subfigure}\hfill
    \begin{subfigure}{0.32\textwidth}
    \includegraphics[width=\linewidth,height=4.5cm]{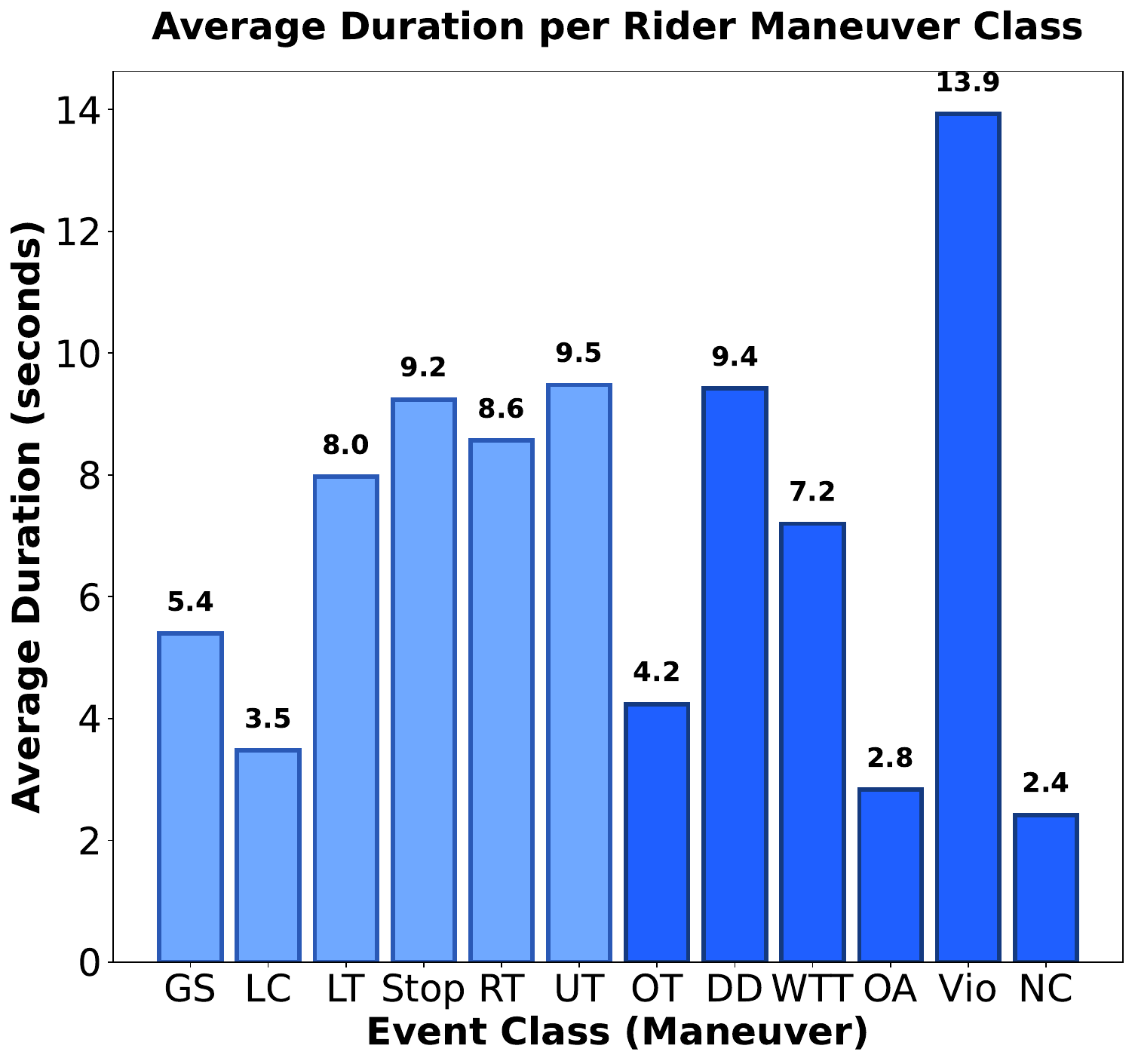}
        \caption{}
        \label{fig:plots_b}
    \end{subfigure}\hfill
    \begin{subfigure}{0.32\textwidth}
        \includegraphics[width=\linewidth,height=4.5cm]{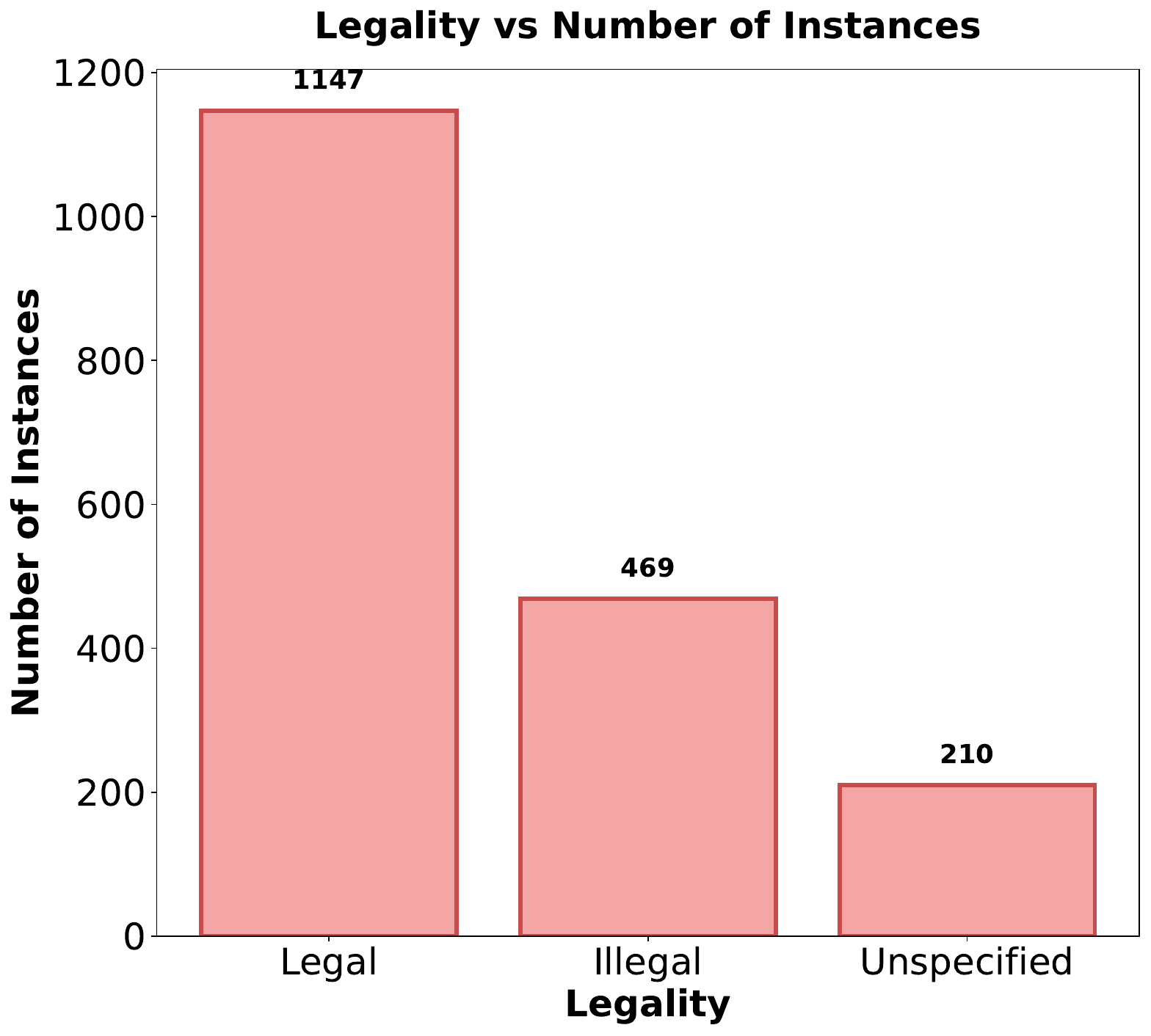}
        \caption{}
        \label{fig:plots_c}
    \end{subfigure}

\begin{subfigure}{0.32\textwidth}
    \includegraphics[width=\linewidth,height=4.5cm]{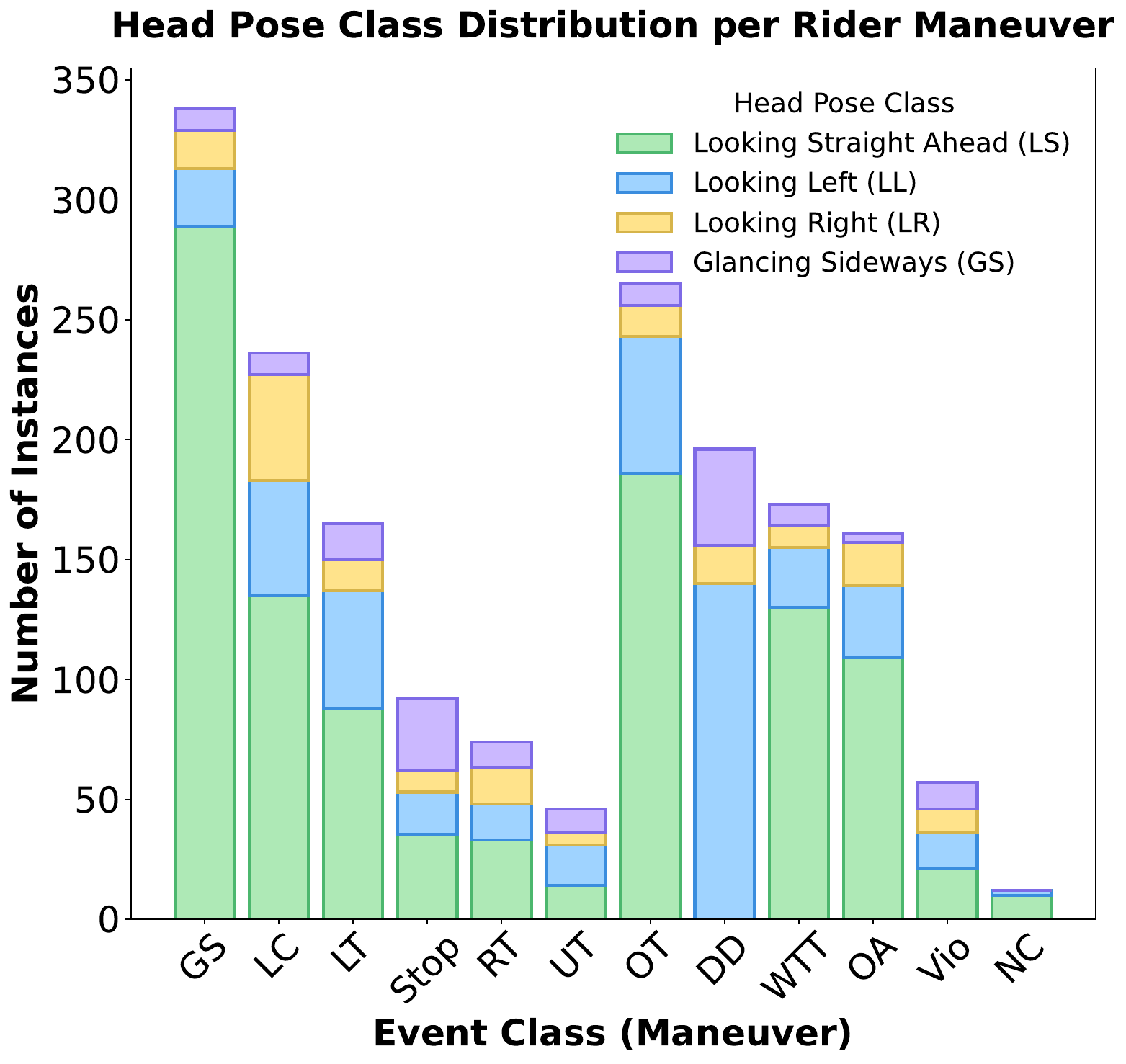}
    \caption{}
    \label{fig:plots_d}
\end{subfigure}\hfill
\begin{subfigure}{0.32\textwidth}
    \includegraphics[width=\linewidth,height=4.5cm]{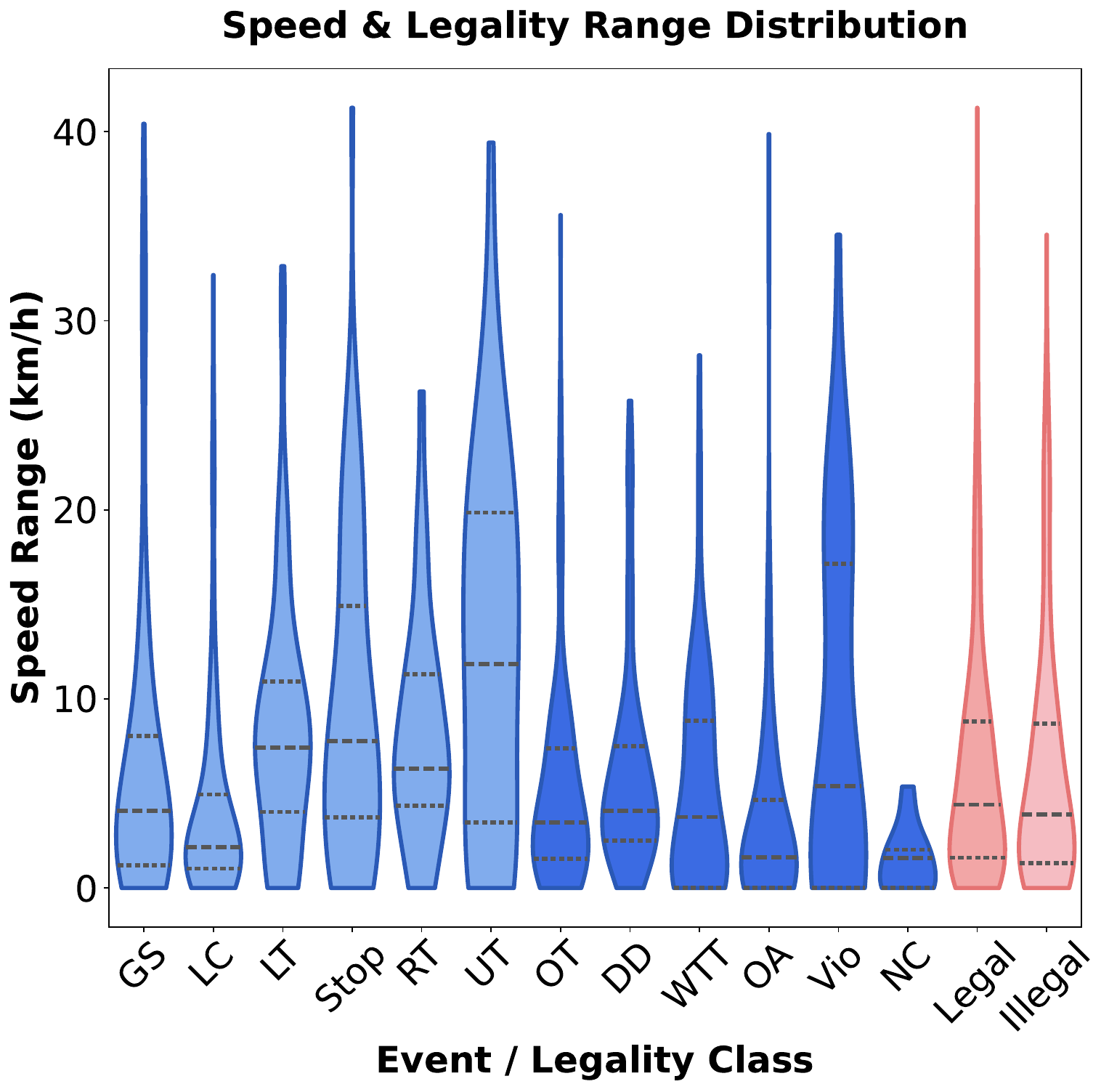}
    \caption{}
    \label{fig:plots_e}
\end{subfigure}\hfill
\begin{subfigure}{0.32\textwidth}
    \includegraphics[width=\linewidth,height=4.5cm]{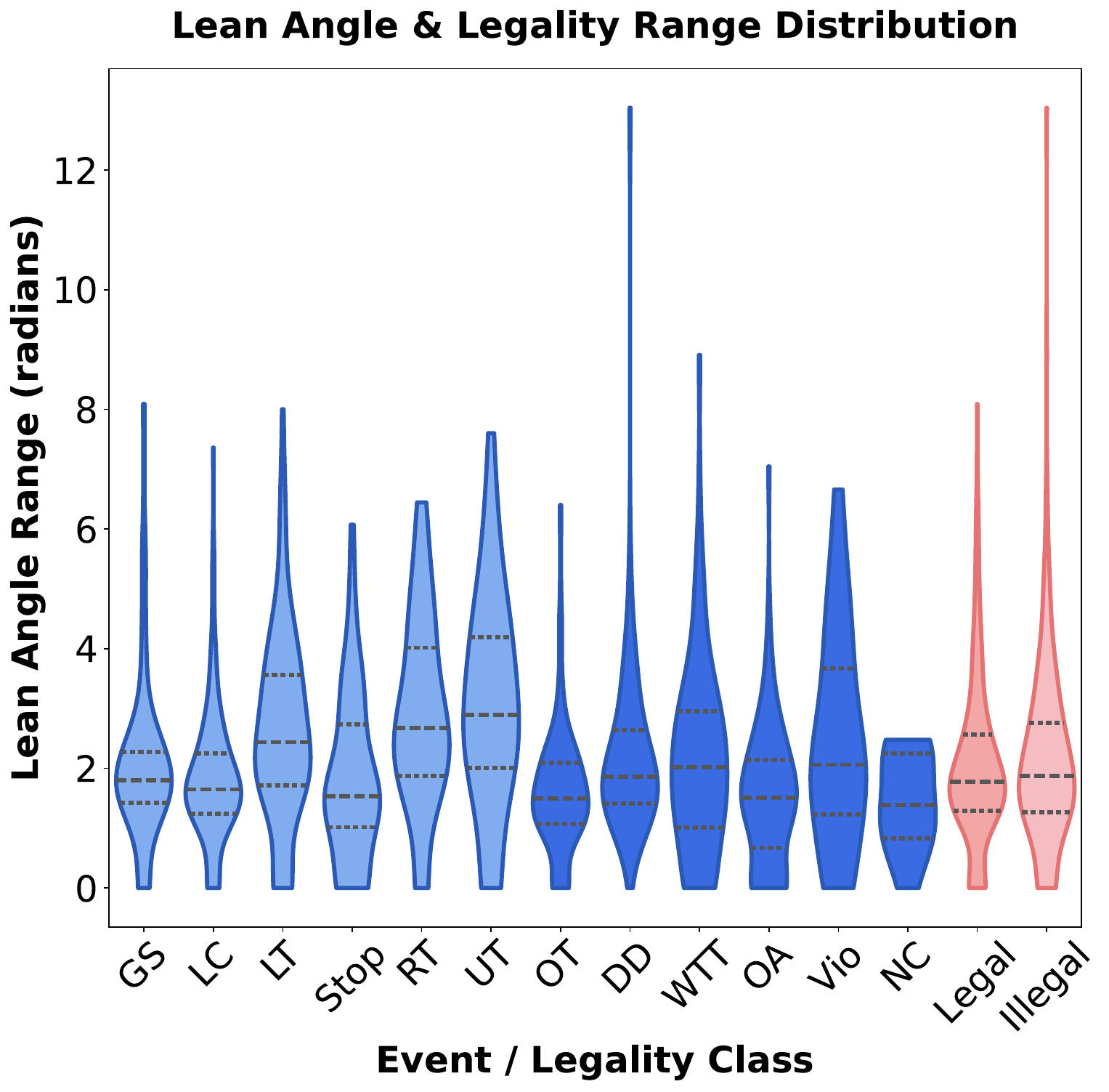}
    \caption{}
    \label{fig:plots_f}
\end{subfigure}

    \caption{\textbf{Annotation Instances and data statistics.} \textit{Top row:} Distribution, average duration and legality of rider maneuvers. \textit{Bottom row:} Distribution of head pose class, speed range, and lean angle per rider maneuver class.}
    \label{fig:six_plots}
\end{figure*}

\noindent \textbf{Data Statistics.}  
Fig.~\ref{fig:six_plots} presents an overview of the MOTOR dataset, which includes 1,629 annotated maneuver instances across 12 classes. Maneuver durations range from 2.4 seconds for near-collisions to nearly 14 seconds for traffic violations. Over 60\% of instances occur in medium/high traffic, and more than 50\% on one- or two-lane city roads (Fig.~\ref{fig:plots_c}), reflecting typical urban conditions for two-wheelers. A unique feature is rider head-pose annotations (Fig.~\ref{fig:plots_d}), showing how attention shifts with traffic context. Speed and lean angle distributions (Figs.~\ref{fig:plots_e}, \ref{fig:plots_f}) reveal strong correlations with maneuvers e.g., high lean during turns and abrupt speed changes in obstruction avoidance or weaving. These statistics highlight the richness of MOTOR in capturing rider state and environmental variability for modeling two-wheeler behavior in dense traffic.

\subsection{Evaluation Tasks and Metrics.} 
We evaluate two tasks on the MOTOR dataset: 
\begin{itemize}
    \item \textbf{Rider Behavior Classification:} Classify rider maneuvers into 11 classes spanning conventional (e.g., turns, lane changes, overtakes) and unconventional (e.g., weaving, obstruction avoidance, violations) behaviors. The \textit{Near Collision} class is excluded due to data sparsity. This task tests a model’s ability to capture diverse and fine-grained two-wheeler actions in dense traffic.
    \item \textbf{Maneuver Legality Classification:} 
    Predict whether a rider maneuver is \textit{Legal}, \textit{Illegal}, or \textit{Unspecified}, going beyond behavior recognition to explicitly assess compliance with traffic rules, crucial for safety-critical and traffic-aware systems.
\end{itemize}
Performance for both tasks was evaluated using \textit{Accuracy} and \(F_1\) scores, reflecting overall correctness and class-balanced performance in skewed label distributions.

\noindent \textbf{Sanity Check.} Two professional annotators, trained on maneuver and traffic violation definitions from~\cite{Indian_Motor_Vehicle_Act}, labeled the dataset under expert supervision. For the first 50 sequences, both annotated independently, after which the expert reviewed their work, resolved discrepancies, and provided feedback. The remaining sequences were also annotated individually, with the expert conducting periodic random checks to ensure accuracy and consistency.  

\noindent \textbf{Ethical Statement}. All participants receive informed consent prior to data collection, with procedures reviewed and approved by the Institutional Research Board (IRB). Riders were over 18 years of age and held valid driving licenses. Data collection was conducted under naturalistic riding conditions, and no rider was instructed to perform any unconventional or unsafe maneuvers. For safety, instructors supervised the process and participants were briefed on risks before each session. The dataset is anonymized and made publicly available strictly for research purposes.

\section{Baselines}
\label{benchmark-sec}
\noindent We design a three stream late-fusion architecture, integrating ego-vehicle frontal-view, rider eye-gaze, and vehicle telemetry, as the baseline for rider behavior and legality classification. The overall pipeline is shown in Fig.~\ref{fig:architecture}. 

\noindent \textbf{Ego-Vehicle Video Stream.} 
The frontal-view video clips are uniformly sampled into 16-frame segments and processed through a video encoder to extract spatio-temporal embeddings. We benchmark four action recognition backbones: CNN-based S3D~\cite{xie2018rethinkingspatiotemporalfeaturelearning} and ResNet3D~\cite{tran2018closerlookspatiotemporalconvolutions}, and transformer-based Video Swin Transformer~\cite{liu2021videoswintransformer} and MViTv2~\cite{li2022mvitv2improvedmultiscalevision}.

\noindent \textbf{Gaze Stream.} 
Rider gaze points, recorded using wearable eye-trackers (Aria~\cite{engel2023projectarianewtool} and Pupil~\cite{kassner2014pupilopensourceplatform}), are used to crop synchronized gaze-centered regions from the video, capturing the rider’s localized attention focus. Similar to the video stream, gaze crops are sampled into 16-frame clips and passed through a video encoder, enabling the model to leverage both global scene dynamics and fine-grained attention cues.

\noindent \textbf{Telemetry Stream.} 
Vehicle telemetry (speed and lean angle) provides a compact yet informative representation of motion dynamics. Each signal is resampled into fixed-length sequences of 64 bins, concatenated into a 128-dimensional vector, and projected into an embedding space using a lightweight three-layer MLP with ReLU activations. This stream captures dynamics that are often difficult to infer directly from video, such as subtle lean angles during turns or speed variations during acceleration and braking.

\noindent \textbf{Late Fusion and Classification.} 
The three embeddings from video, gaze, and telemetry are concatenated in a late-fusion scheme to form a unified representation. This fused vector is passed through an MLP classifier consisting of LayerNorm, two hidden layers with ReLU activations and dropout, and a final linear layer, which outputs either the rider behavior or legality classes.

\noindent \textbf{Loss Function.} 
Rider behavior and legality classification face strong class imbalance, dominated by common maneuvers such as \textit{going straight} or \textit{stopping}. We use focal loss~\cite{lin2017focal} with class-balanced weights to address this. For a sample with ground-truth class \(y\) and predicted probability \(p_y\), the focal loss is defined as:
\[
\mathcal{L}_{\text{focal}} = - \alpha_y (1 - p_y)^{\gamma} \log(p_y),
\]

where \(\alpha_y\) adjusts for class frequency and \(\gamma\) controls the focusing parameter. This formulation emphasizes on the minority maneuvers, such as weaving through traffic.

\section{Experiments}
\subsection{Experimental Settings}
\label{sec:implementation}
\noindent \textbf{Implementation Details.} 
All experiments were conducted on a server equipped with an Intel Xeon E5-2640 v4 CPU and four NVIDIA RTX 2080 Ti GPUs. Each of the four video backbones, i.e., S3D~\cite{xie2018rethinkingspatiotemporalfeaturelearning}, ResNet-3D~\cite{tran2018closerlookspatiotemporalconvolutions}, Video Swin Transformer~\cite{liu2021videoswintransformer}, MViTv2~\cite{li2022mvitv2improvedmultiscalevision} were evaluated under identical conditions to ensure fairness of comparison.

\noindent \textbf{Sampling Strategy.} 
For each video clip, 16 RGB frames were uniformly sampled. Rider gaze crops were synchronized at the same rate to maintain temporal alignment. Telemetry signals were discretized into 64-bin sequences each and concatenated into a 128-dimensional vector per clip, ensuring consistent representation.

\noindent \textbf{Training Procedure.} 
All video backbones were initialized with official PyTorch~\cite{paszke2019pytorch} pretrained weights. All but the final block of each backbone were frozen to reduce computation while enabling task-specific adaptation. The unfrozen block, telemetry encoder and fusion head, were trained end-to-end for 50 epochs with a batch size of 8. We used the AdamW optimizer with cosine annealing, setting learning rates to \(1 \times 10^{-4}\) for the telemetry encoder and fusion head, and \(3 \times 10^{-5}\) 
for the unfrozen video block, with a weight decay of \(1 \times 10^{-4}\). Different learning rates were chosen to allow faster convergence for newly initialized modules.

\subsection{Rider Behavior Classification Results}
\noindent We compare the CNN-based (S3D~\cite{xie2018rethinkingspatiotemporalfeaturelearning}, ResNet3D~\cite{tran2018closerlookspatiotemporalconvolutions}) and transformer-based (MViTv2~\cite{li2022mvitv2improvedmultiscalevision}, Video Swin Transformer (SwinT)~\cite{liu2021videoswintransformer}) baselines on our MOTOR dataset as reported in Table~\ref{tab:maneuver_classification}. We observe that the transformer-based SwinT~\cite{liu2021videoswintransformer} baseline outperforms accuracy and $F_1$ score using all data modalities consistently. SwinT gains over MViTv2 by a margin of 11.4\% accuracy and 14.0\% $F_1$ score, showing its effectiveness in modeling complex rider behaviors. Within CNNs, ResNet3D emerges as the stronger architecture, outperforming S3D across all modality settings. A possible reason why MViTv2 under performs compared to ResNet3D and SwinT is its sensitivity to dataset scale and the noise due the the camera movements and vibrations, while the SwinT's hierarchical attention makes it more robust and better suited to dense, noisy scenarios.
Table~\ref{tab:maneuver_classification} also reports accuracy and $F_1$ score across different modalities, underscoring the importance of gaze and telemetry for behavior classification. Using SwinT with all modalities, we achieve accuracy of 52.9\%. Removing gaze reduces accuracy by 1.6\%, while removing telemetry leads to a 2.6\% drop in accuracy. Excluding both causes a 5.2\% decline in accuracy, highlighting the effectiveness of full fusion (RGB+Gaze+Telemetry). These results demonstrate that combining attention (gaze) and kinematic (telemetry) cues with visual information significantly improves rider behavior prediction.

\begin{table}[H]
\centering
\caption{\textbf{Rider Behavior Classification.} Comparison of CNN and Transformer-based baselines on MOTOR dataset across different modality combinations.}
\resizebox{\columnwidth}{!}{
\begin{tabular}{l l c c c}
\toprule
\textbf{Baseline} & \textbf{Data Modalities} & \textbf{ACC (↑)} & \textbf{$F_1$ (↑)} & \textbf{Params (M) (↓)} \\
\midrule
\multicolumn{5}{l}{\textit{CNN-based Backbones}} \\
\midrule
S3D~\cite{xie2018rethinkingspatiotemporalfeaturelearning} & RGB & 38.3 & 35.3 & 2.4 \\
 & RGB+Gaze & 37.3 & 34.2 & 4.7 \\
 & RGB+Telemetry & 39.2 & 35.8 & 2.5 \\
 & RGB+Gaze+Telemetry & 39.3 & 34.2 & 4.85 \\
\midrule
ResNet3D~\cite{tran2018closerlookspatiotemporalconvolutions} & RGB & 48.7 & 45.4 & 14.0 \\
 & RGB+Gaze & 48.2 & 47.2 & 28.0 \\
 & RGB+Telemetry & 48.8 & 47.1 & 14.1 \\
 & RGB+Gaze+Telemetry & 49.1 & 48.1 & 28.5 \\
\midrule
\multicolumn{5}{l}{\textit{Transformer-based Backbones}} \\
\midrule
MViTv2~\cite{li2022mvitv2improvedmultiscalevision} & RGB & 32.6 & 32.4 & 7.5 \\
 & RGB+Gaze & 39.4 & 34.5 & 15.01 \\
 & RGB+Telemetry & 39.8 & 36.1 & 7.6 \\
 & RGB+Gaze+Telemetry & 41.5 & 37.5 & 15.1 \\
\midrule
Swin T~\cite{liu2021videoswintransformer} & RGB & 47.7 & 46.3 & 7.6 \\
 & RGB+Gaze & 50.3 & 46.9 & 15.1 \\
 & RGB+Telemetry & 51.3 & 47.2 & 7.7 \\
 & \textbf{RGB+Gaze+Telemetry} & \textbf{52.9} & \textbf{51.5} & 15.2 \\
\bottomrule
\end{tabular}
}
\label{tab:maneuver_classification}
\end{table}


\begin{figure*}[t] 
  \centering
  \includegraphics[width=\textwidth]{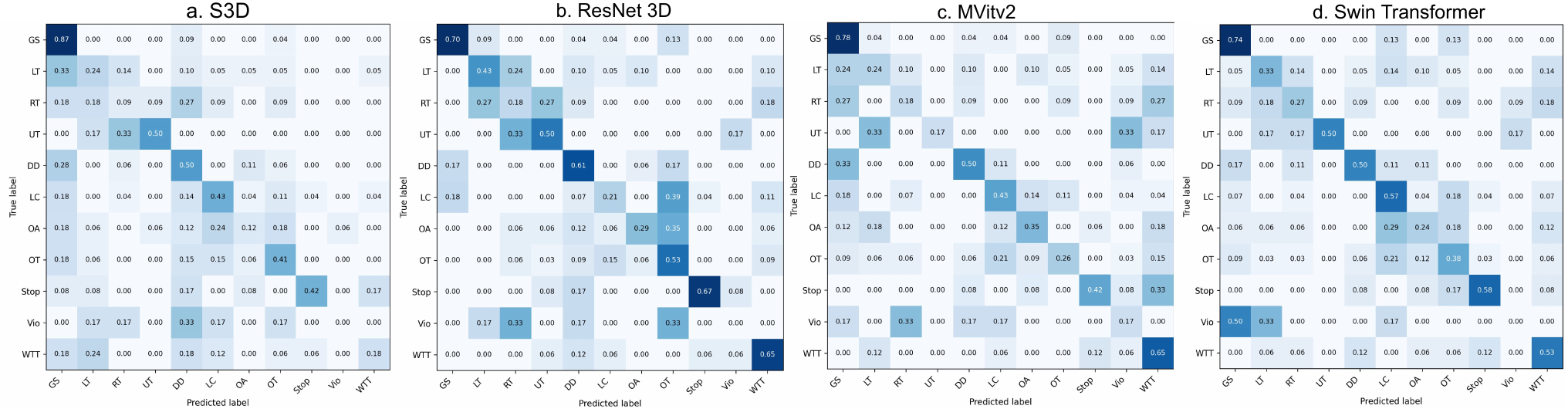}
  \caption{Confusion matrices for rider behavior classification using S3D~\cite{xie2018rethinkingspatiotemporalfeaturelearning}, ResNet3D~\cite{tran2018closerlookspatiotemporalconvolutions}, MViTv2~\cite{li2022mvitv2improvedmultiscalevision}, and SwinT~\cite{liu2021videoswintransformer} on the MOTOR dataset.}

  \label{fig:confusion_matrices}
\end{figure*}

\vspace{-0.4cm}
\subsection{Maneuver Legality Classification Results}


\begin{table}[t]
\centering
\caption{\textbf{Rider Legality Classification:} CNN and Transformer-based baselines on MOTOR dataset across different modality combinations.}
\resizebox{\columnwidth}{!}{
\begin{tabular}{l l c c c}
\toprule
\textbf{Baseline} & \textbf{Data Modalities} & \textbf{ACC (↑)} & \textbf{$F_1$ (↑)} & \textbf{Params (M) (↓)} \\
\midrule
\multicolumn{5}{l}{\textit{CNN-based Backbones}} \\
\midrule
S3D~\cite{xie2018rethinkingspatiotemporalfeaturelearning} & RGB & 62.9 & 48.2 & 2.4 \\
 & RGB+Gaze & 62.4 & 48.8 & 4.7 \\
 & RGB+Telemetry & 64.5 & 47.8 & 2.5 \\
 & RGB+Gaze+Telemetry & 64.9 & 51.3 & 4.8 \\
\midrule
ResNet3D~\cite{tran2018closerlookspatiotemporalconvolutions} & RGB & 59.6 & 45.1 & 14.0 \\
 & RGB+Gaze & 60.3 & 45.7 & 28.0 \\
 & RGB+Telemetry & 61.8 & 46.9 & 14.1 \\
 & RGB+Gaze+Telemetry & 62.9 & 47.7 & 28.5 \\
\midrule
\multicolumn{5}{l}{\textit{Transformer-based Backbones}} \\
\midrule
MViTv2~\cite{li2022mvitv2improvedmultiscalevision} & RGB & 58.2 & 45.8 & 7.5 \\
 & RGB+Gaze & 61.9 & 46.2 & 15.0 \\
 & RGB+Telemetry & 62.6 & 49.4 & 7.6 \\
 & RGB+Gaze+Telemetry & 64.3 & 52.1 & 15.1 \\
\midrule
Swin T~\cite{liu2021videoswintransformer} & RGB & 58.4 & 47.9 & 7.6 \\
 & RGB + Gaze & 62.7 & 48.5 & 15.1 \\
 & RGB + Telemetry & 65.0 & 53.5 & 7.7 \\
 & \textbf{RGB+Gaze+Telemetry} & \textbf{69.0} & \textbf{53.6} & 15.2 \\
\bottomrule
\end{tabular}
}
\label{tab:legality_classification}
\end{table}


\noindent We also benchmark CNN-based models and transformer-based baselines for maneuver legality classification on the MOTOR dataset as shown in Table~\ref{tab:legality_classification}. Among all baselines, SwinT~\cite{liu2021videoswintransformer} achieves the best overall performance, reaching 69.0\% accuracy and 53.6\% $F_1$ score with all modalities. Compared to MViTv2, SwinT improves by 4.7\% in accuracy, demonstrating stronger robustness in noisy, heterogeneous traffic conditions. Interestingly, within CNNs, S3D proves to be highly competitive: despite having fewer trainable parameters, it outperforms ResNet3D across all modality settings (by 2.1\% accuracy and 3.6\% $F_1$ score), and even approaches SwinT's transformer-level performance. This may be attributed to S3D’s lightweight 3D convolutions, which capture short-term motion cues crucial for legality assessment, such as detecting wrong-lane usage, illegal turns, or overtakes from the wrong side. Table~\ref{tab:legality_classification} also highlights the contribution of different modalities. Full fusion on SwinT yields the strongest performance, while removing gaze reduces accuracy by 4.0\% and excluding telemetry leads to a 6.3\% drop. Eliminating both modalities results in a 10.6\% decline compared to full fusion. These results further highlight the role of gaze as a complementary attention cue and telemetry as a dominant kinematic signal (e.g., speed and lean angle) in shaping accurate legality prediction for two-wheelers and to understand the two-wheeler behavior.

\subsection{Analysis of Confusion Matrices for Rider Maneuver Classification}

\noindent Fig.~\ref{fig:confusion_matrices} presents the confusion matrices for rider maneuver classification across backbones with gaze and telemetry fusion. Among them, the SwinT delivers the most reliable predictions, showing clear separation between \textit{lane change} and \textit{overtake}, \textit{stop} and \textit{going straight}, as well as \textit{left} versus \textit{right turns}. This suggests that its hierarchical attention effectively captures temporal dependencies and contextual cues in dense traffic, making it more robust to noise and ambiguous rider motion patterns.
Nevertheless, some confusions persist. \textit{Obstruction avoidance} is frequently misclassified as \textit{lane change} or \textit{overtake}, since all involve lateral deviations. Similarly, \textit{traffic violations} are often misclassified because their recognition depends on scene-level cues (e.g., traffic signals, road markings) that are not always visible in dense, unstructured traffic conditions.

\vspace{-0.1cm}
\section{Conclusion and Future Work}
\noindent In this work, we presented \textbf{MOTOR}, the first large-scale, multi-rider, multi-view, and multimodal dataset dedicated to understanding two-wheeler behavior in dense, unstructured traffic. MOTOR integrates synchronized video, rider gaze, audio, and telemetry with rich annotations covering conventional and unconventional maneuvers, along with legality labels. Through extensive experiments and analysis we highlight some key lessons for two-wheeler behavior: gaze provides complementary attention cues, telemetry captures dominant kinematic patterns such as lean and speed, and unconventional behaviors remain challenging to classify due to their variability and overlap with conventional maneuvers. Beyond benchmarking, MOTOR offers a valuable resource for the research community to explore legality-aware modeling and the development of safety-critical applications tailored to two-wheelers. In future, we aim to expand on the dataset with diverse weather and lighting conditions as well as look to improve the baseline architecture by fusing more camera streams and modalities. 

\vspace{-0.2cm}
\section{Acknowledgment}
\noindent The project was supported by iHub-Data and Mobility at IIIT Hyderabad. We would also like to thank Shubham Kumar and Md. Azhar for their help with the data collection.

\bibliographystyle{IEEEtran}
\vspace{-.25cm}
\bibliography{reference}

\end{document}